\documentclass{article}

\usepackage{PRIMEarxiv}

\usepackage[utf8]{inputenc} 
\usepackage[T1]{fontenc}    
\usepackage{hyperref}       
\usepackage{url}            
\usepackage{booktabs}       
\usepackage{amsfonts}       
\usepackage{nicefrac}       
\usepackage{amsmath,amssymb,amsthm}
\usepackage{microtype}      
\usepackage{lipsum}
\usepackage{fancyhdr}       
\usepackage{graphicx}       
\graphicspath{{media/}}     

\title{Synaptic motor adaptation: A three-factor learning rule for adaptive robotic control in spiking neural networks
}

\author{
  Samuel Schmidgall$^{1,2}$, Joe Hays$^{1}$ \\
  $^{1}$U.S. Naval Research Laboratory, $^{2}$Johns Hopkins University \\
  \texttt{\{samuel.schmidgall, joe.hays\}email@email} \\
}

\begin{document}
\maketitle

\begin{abstract}
Legged robots operating in real-world environments must possess the ability to rapidly adapt to unexpected conditions, such as changing terrains and varying payloads. This paper introduces the Synaptic Motor Adaptation (SMA) algorithm, a novel approach to achieving real-time online adaptation in quadruped robots through the utilization of neuroscience-derived rules of synaptic plasticity with three-factor learning. To facilitate rapid adaptation, we meta-optimize a three-factor learning rule via gradient descent to adapt to uncertainty by approximating an embedding produced by privileged information using only locally accessible onboard sensing data. Our algorithm performs similarly to state-of-the-art motor adaptation algorithms and presents a clear path toward achieving adaptive robotics with neuromorphic hardware.
\end{abstract}

\keywords{robot learning, spiking neural network, synaptic plasticity, neuromodulation, online learning}

\section{Introduction}

Legged robots have made significant progress in the last four decades using physical dynamics modeling and control theory, requiring considerable expertise from the designer \cite{raibert1986legged, raibert2008bigdog, feng2014optimization, kuindersma2016optimization}. In recent years, researchers have shown interest in using reinforcement and imitation learning techniques to reduce the designer's burden and enhance performance \cite{yang2020data, lee2020learning, rudin2022learning}. However, adaptation to new domains has remained a challenging problem due to various factors such as the differences in data distribution between the source and target domains, as well as the inherent complexity of the underlying relationships between the input and output variables (i.e. dynamic system uncertainties), which often necessitate significant modifications to the learning algorithms and architectures in order to achieve satisfactory results in the target domain \cite{hofer2020perspectives}. 

Neuromorphic computing offers a promising approach to address the challenges of adaptation in legged robotics by enabling the development of more efficient and adaptive algorithms that can better emulate the neural structures and functions of biological systems. In addition, these systems are extremely energy efficient \cite{painkras2013spinnaker, DBLP:journals/corr/EsserMACAABMMBN16, davies2018loihi}, enabling robotic learning algorithms to operate across long timescales without recharging. Many neuromorphic chips are betting on local learning rules, such as Hebbian and spike-timing dependent plasticity rules, to provide \textit{on-chip learning} for efficient learning on edge-computing applications \cite{davies2018loihi, pehle2022brainscales, painkras2013spinnaker, jin2010implementing}. Local learning rules offer several advantages beyond their biological inspiration, including computational efficiency, scalability, and the ability to adapt to dynamic environments. Unlike traditional machine learning algorithms that require large amounts of training data and significant computational resources, local learning rules can learn from small amounts of data and adapt in real-time, making them particularly useful for applications in edge computing \cite{vertechi2014unsupervised, kaiser2020synaptic, wu2022brain}. Furthermore, since the learning is distributed across the network of neurons, local learning rules are highly parallelizable, allowing for efficient processing of large amounts of data. 

Recently, there has been notable progress in developing algorithms that employ local learning rules, due to the advancements in the theory of three-factor learning in neuroscience \cite{fremaux2016neuromodulated, gerstner2018eligibility}. This theory offers a solution for assigning credit to synapses over time without relying on the backpropagation of errors, which is typically used for credit assignment in machine learning applications. The current most effective local learning rules for neuromorphic devices are based on this theory, and show promising potential for enabling on-chip learning in various real-world applications \cite{bellec2020solution, schmidgall2021spikepropamine, wu2022brain}. 

\begin{figure*}
  \includegraphics[width=\textwidth]{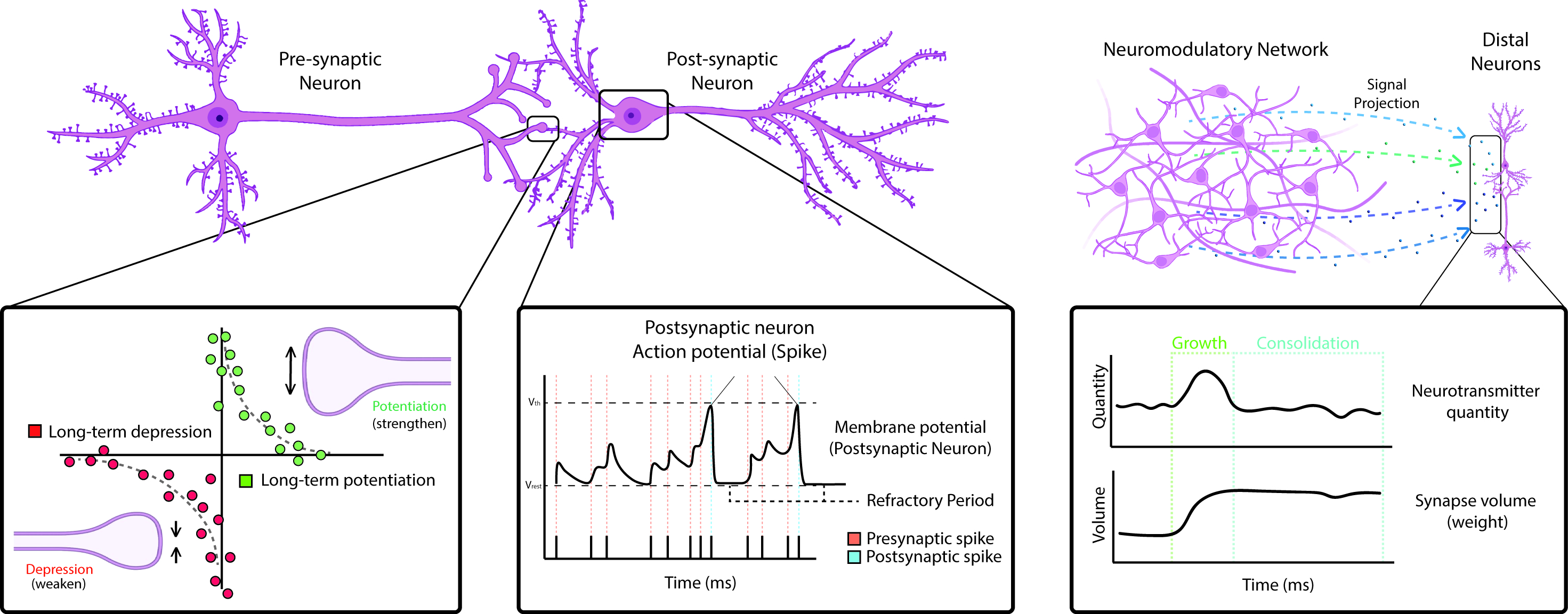}
  \caption{Graphical description of three-factor learning. (Left) Long-term potentiation and depression based on pre- and post-synaptic spike timings. (Middle) Membrane potential dynamics of the post-synaptic neuron. (Right) Neuromodulator quantity affects the growth and consolidation of the synaptic weights.}
  \label{fig:teaser}
\end{figure*}

Independently and in parallel, significant strides have been made toward developing adaptive robotic controllers for legged robots. These methods, termed \textit{motor adaptation} (MA) algorithms, learn how to estimate their current environmental factors (e.g. friction coefficients, terrain, etc) from locally accessible data, which is provided as state input into the network \cite{kumar2021rma, kumar2022adapting, agarwal2023legged, qi2023hand, fu2023deep}. In this work, we introduce a motor adaptation which uses neuroscience-derived rules of plasticity together with a third factor signal to dynamically update the \textit{synaptic weights} of the network. This method, called Synaptic Motor Adaptation (SMA), provides a novel approach to motor adaptation by enabling the policy to learn from new experiences in real-time rather than simply updating its state input. 

SMA is particularly well-suited to legged robots as it allows the network to update its connections with respect to the current environment conditions, such as uneven terrain, while maintaining stable control over the robot. The proposed three-factor learning rule used in SMA build on the work of differentiable plasticity \cite{schmidgall2021spikepropamine, schmidgall2022learning, schmidgall2023metaspikepropamine}, which makes it amenable to gradient descent optimization. This approach has the potential to significantly improve the performance and adaptability of legged robots, which could have wide-ranging applications in  the field of robotics, particularly for the deployment of neuromorphic devices.

\section{Background \& Related work}

\begin{figure*}
    \centering
\includegraphics[width=0.98\linewidth]{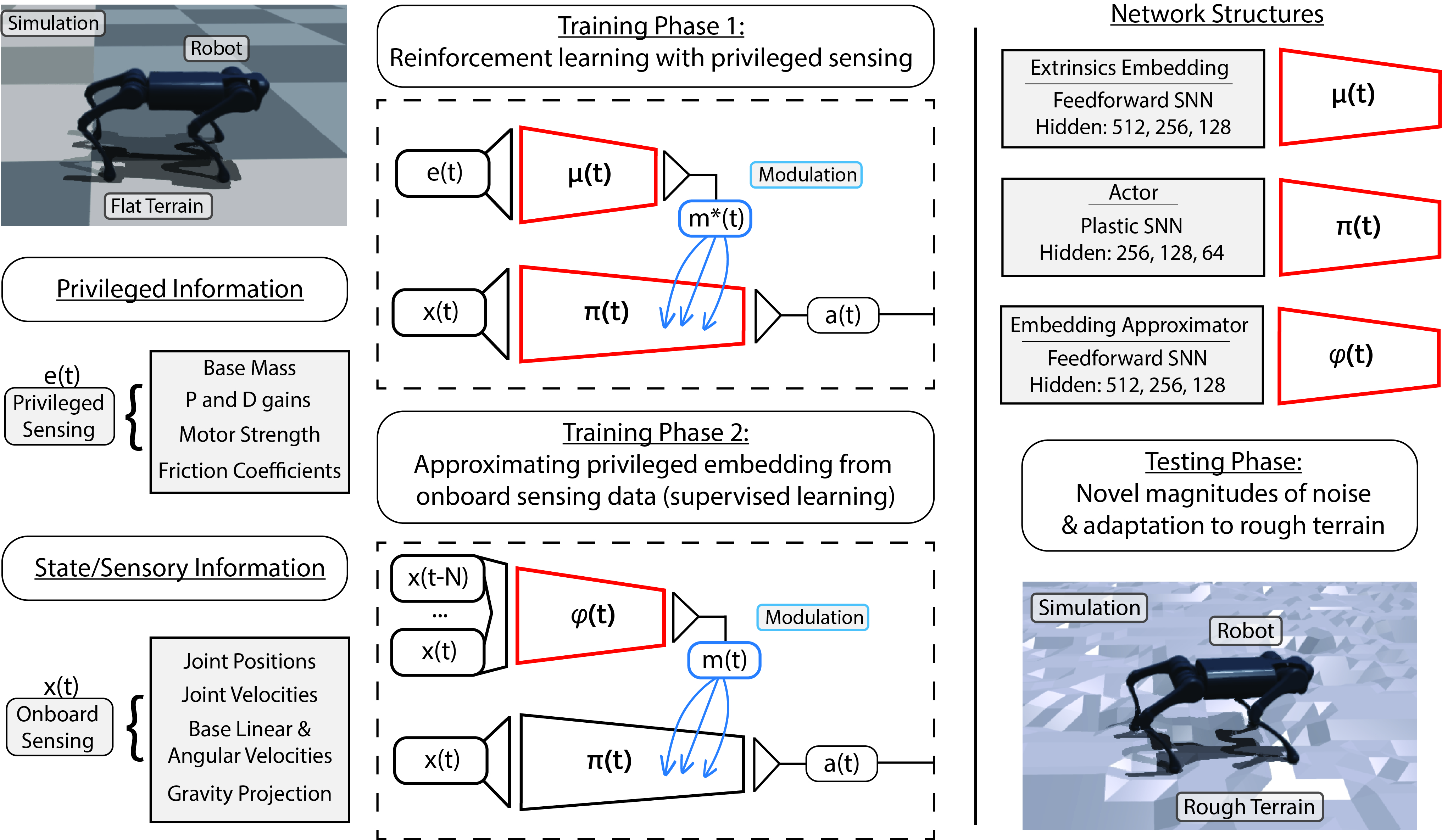}
\caption{\textbf{Overview of synaptic motor adaptation algorithm}. (Left) Privileged and sensory information vectors. (Middle) Training phase 1 consists of reinforcement leanring with privileged sensing with the neuromodulatory extrinsics embedding $\mu(t)$ and the actor $\pi(t)$. Phase 2 consists of approximating the dynamics of the neuromodulatory extrinsics embedding. (Right) Description of the network structures of each model used. (Right bottom) Image of robot on rough terrain, not encountered during training period.}
\end{figure*}

\subsection*{Motor Adaptation Algorithms}

Robotic learning has remained a major challenge in AI since successful deployment would require the algorithm to adapt in real-time to unseen situations, such as dynamic payloads, novel terrain dynamics, as well as hardware degradation over time. This problem has remained a major hurdle since the majority of deep learning algorithms would train a network in simulation \textit{offline} and then fix the network weights for \textit{online} deployment. Significant advancements have been realized recently with the introduction of \textit{motor adaptation} algorithms \cite{kumar2021rma, kumar2022adapting, agarwal2023legged, qi2023hand, fu2023deep}, which act much like a system identification estimator, with the difference that (1) the estimate is a learned embedding containing only the most vital information for adaptation rather than the entirety of the system dynamics and (2) the estimate is made very rapidly from a temporal history of sensory information.

Motor adaptation algorithms typically consist of two components: a base policy $\pi$ and an environment factor encoder $\mu$. During the first phase of simulated training, the factor encoder $\mu$ takes as input privileged information from the environment $e(t)$ that would not be accessible to a deployed system (e.g. friction, motor strength, robot center of mass) and produces a low-dimensional output embedding $z(t)$ referred to as an latent extrinsic vector. The latent vector $z(t)$ is then provided as input to the base policy $\pi$ and optimized by the base policy loss such that $z(t)$ proves a useful latent representation for $\pi$ so that it can better solve its objective. This process can be described by the following equations:

\begin{equation}\label{eq:FactorEncoder}
    z(t) = \mu(e(t))
\end{equation}
\begin{equation}\label{eq:PiModule}
    a(t)= \pi(x(t), a(t-1), z(t)).
\vspace{2mm}
\end{equation}

In the second phase of training, an environment factor \textit{estimator} $\phi$ is trained via regression to match the output of the environment factor encoder $\mu$ using a time history of state and action pairs ($a(t-N-1),s(t-N),..., a(t-1), s(t)$). In essence, an online approximation of the extrinsics embedding $z(t)$ is generated using information accessible to the robot.

\begin{equation}
    \hat{z}(t) = \phi(a(t-N-1),s(t-N),..., a(t-1), s(t))
    \label{eq:embedapprox}
\end{equation}
\begin{equation}
    a(t) = \pi(x(t), a(t-1), \hat{z}(t)).
\vspace{2mm}
\end{equation}

Motor adaptation algorithm have been demonstrated to significantly improve the adaptation abilities of robotic learning in simulation, and has also been used for the deployment of networks trained entirely in simulation to robotic hardware (sim2real) \cite{kumar2021rma, kumar2022adapting, agarwal2023legged, qi2023hand, fu2023deep}.

\subsection*{Synaptic plasticity and three-factor learning}

Plasticity in the brain refers to the capacity of experience to modify the function of neural circuits. The plasticity of synapses refers to the modification of the strength of synaptic transmission based on local activity and is currently the most widely investigated mechanism by which the brain adapts to new information \cite{citri2008synaptic, abraham2019plasticity}. Methods in deep learning are based on changing weights from experience, typically through the use of the algorithm \textit{backpropagation,} which makes predictions based on input and uses the chain rule to back-propagate errors through the network \cite{rumelhart1986learning}. While there are parallels between backpropagation and synaptic plasticity, there are many significant ways in which they differ in operation compared to the brain \cite{lillicrap2020backpropagation}. Three-factor learning rules have been proposed as a much more plausible theoretical framework for understanding how meaningful changes are made in the brain \cite{fremaux2016neuromodulated, gerstner2018eligibility}. Below, we introduce a pair-based model of plasticity and the theory of three-factor learning.

\subsubsection*{Pair-based spike-timing dependent plasticity} The pair-based spike-timing dependent (STDP) model is a plasticity rule that governs changes in synapses based on the timing relationship between pairs of pre- and post-synaptic spikes \cite{caporale2008spike}. This model was derived from experiments which observed that the precise timing of spikes can describe synaptic long-term potentiation (LTP, increase in weight) and long-term depression (LTD, decrease in weight).

We begin by describing the timing dynamics of pre- and post-synaptic spikes through an iterative update rule, referred to as a \textit{synaptic trace} (\textit{also see Figure 1}):

\begin{equation}\label{eq:STDPVariableX}
\textbf{\textit{x}}_{i}^{(l)}(t+\Delta\tau) = \alpha_{x} \textbf{\textit{x}}_{i}^{(l)}(t) + f(\textbf{\textit{x}}_{i}^{(l)}(t)) \textbf{\textit{s}}^{(l)}_{i}(t).
\vspace{1.3mm}
\end{equation}

The precise physiological interpretation of the activity trace $\textbf{\textit{x}}_{i}^{(l)}(t) \in \mathbb{R} > 0$ is not well-defined, as there are several possible representations for this activity. In the case of pre-synaptic events, it could correspond to the quantity of bound glutamate or the number of activated NMDA receptors, while for post-synaptic events it could reflect the synaptic voltage generated by a backpropagating action potential or the amount of calcium influx through a backpropagating action potential \cite{gerstner2014neuronalD}.

The activity trace $\textbf{\textit{x}}_{i}^{(l)}(t)$ is reduced to zero with the variable $\alpha_{x} \in (0, 1)$, where $\alpha_{x}$ is commonly expressed as $(1-\frac{1}{\tau})$ and decays at a rate dependent on the time constant $\tau \in \mathbb{R} > 1$. The update of the synaptic trace is determined by a function $f: \mathbb{R} \to \mathbb{R}$, which is proportional to the presence of a spike $\textbf{\textit{s}}^{(l)}_{i}(t)$. This all-to-all synaptic trace scheme pairs each pre-synaptic spike with every post-synaptic spike indirectly via the decaying trace. In the linear update rule, which is used in this work, the trace is updated by a constant factor $\boldsymbol\beta$ when a spike $\textbf{\textit{s}}^{(l)}_{i}(t)$ occurs.

\begin{equation}\label{eq:STDPVariableLin}
\textbf{\textit{x}}_{i}^{(l)}(t+\Delta\tau) = \alpha_{x} \textbf{\textit{x}}_{i}^{(l)}(t) +  \boldsymbol\beta\textbf{\textit{s}}^{(l)}_{i}(t).
\vspace{1.3mm}
\end{equation}

Next, we describe the pair-based STDP rule, which describes LTP (left-hand side of Equation \ref{eq:PairbasedSTDPDelta}) and LTD (right-hand side of Equation \ref{eq:PairbasedSTDPDelta}) via pairs of spikes and synaptic traces:

\begin{equation}\label{eq:PairbasedSTDP}
\textbf{\textit{W}}_{i,j}^{(l)}(t+\Delta\tau) = \textbf{\textit{W}}_{i,j}^{(l)}(t) + \Delta_{\textbf{\textit{W}}}(t)
\end{equation}
\begin{equation}\label{eq:PairbasedSTDPDelta}
\Delta_{\textbf{\textit{W}}}(t) = \textbf{\textit{A}}_{+,i,j}\textbf{\textit{x}}_{i}^{(l-1)}(t) \textbf{\textit{s}}_{j}^{(l)}(t) - \textbf{\textit{A}}_{-,i,j}\textbf{\textit{x}}_{j}^{(l)}(t) \textbf{\textit{s}}_{i}^{(l-1)}(t).
\vspace{2mm}
\end{equation}

When a post-synaptic firing occurs ($\textbf{\textit{s}}_{j}^{(l)}(t) = 1$), weight potentiation occurs by a quantity proportional to the pre-synaptic trace ($\textbf{\textit{x}}_{i}^{(l-1)}(t)$). Similarly, when a pre-synaptic firing occurs ($\textbf{\textit{s}}_{i}^{(l-1)}(t) = 1$), weight depression occurs by a quantity proportional to the post-synaptic trace ($\textbf{\textit{x}}_{j}^{(l)}(t)$). Potentiation and depression are scaled by constants $\textbf{\textit{A}}_{+,i,j} \in \mathbb{R}$ and $\textbf{\textit{A}}_{-,i,j} \in \mathbb{R}$, respectively, which characterize the rate of change of LTP and LTD. Typically, Hebbian pair-based STDP models define $\textbf{\textit{A}}_{+,i,j} > 0$ and $\textbf{\textit{A}}_{-,i,j} > 0$, while anti-Hebbian models define $\textbf{\textit{A}}_{+,i,j} < 0$ and $\textbf{\textit{A}}_{-,i,j} < 0$. We initialize our learning rule to be \textit{Hebbian}, but do not constrain the optimization, thus allowing our initially Hebbian rule to become anti-Hebbian or any other variations of the pair-based STDP rule.


\subsubsection*{Eligibility traces and three-factor plasticity}

Rather than directly modifying the synaptic weight, local synaptic activity leaves an activity flag, or eligibility trace, at the synapse \cite{gerstner2018eligibility}. The eligibility trace does not immediately produce a change, rather, weight change is realized in the presence of an additional signal, which is discussed below. In a Hebbian learning rule, the eligibility trace can be described by the following equation:

\begin{equation}\label{eq:Eligibility}
\textbf{\textit{E}}_{i,j}^{(l)}(t+\Delta\tau) = \gamma \textbf{\textit{E}}_{i,j}^{(l)}(t) + \boldsymbol\alpha_{i,j}f_{i}(x^{(l-1)}_{i})g_{j}(x^{(l)}_{j}).
\end{equation}

The decay rate of the trace is determined by the constant $\gamma \in [0, 1]$, where a higher value of $\gamma$ results in a faster decay. The constant $\boldsymbol\alpha_{i,j} \in \mathbb{R}$ determines the rate at which activity trace information is incorporated into the eligibility trace. The functions $f_{i}$ and $g_{j}$ depend on the pre- and post-synaptic activity traces, $x^{(l-1)}_{i}$ and $x^{(l)}_{j}$, respectively. These functions are indexed by the corresponding pre- and post-synaptic neuron, $i$ and $j$, as the eligibility dynamics of synaptic activity may be influenced by neuron type or the region of the network.

Theoretical neuroscience literature suggests that eligibility traces alone cannot bring about a change in synaptic efficacy \cite{gerstner2018eligibility, fremaux2016neuromodulated}. Rather, weight changes require the presence of a third signal.

\begin{equation}\label{eq:ModEligibilityNeuronMod}
\textbf{\textit{W}}_{i,j}^{(l)}(t+\Delta\tau) = \textbf{\textit{W}}_{i,j}^{(l)}(t) + M_{j}(t) \textbf{\textit{E}}_{i,j}^{(l)}(t).
\vspace{2mm}
\end{equation}

Here, $M_j(t) \in \mathbb{R}$ is a regional \textit{\textbf{third factor}} known as a neuromodulator, acting as an abstract representation of a biological process. Without the presence of the neuromodulatory signal ($M_j(t) = 0$), weight changes do not occur. In the presence of certain stimuli, the magnitude and direction of change in $M_j(t)$ determine both long-term potentiation (LTP) and long-term depression (LTD), causing them to scale and reverse. Three-factor learning rules are powerful in their descriptive capabilities, and have been used to describe approximations to Backpropagation Through Time (BPTT) \cite{bellec2019biologically, bellec2020solution} and Bayesian inference \cite{aitchison2021synaptic}.

\section{Synaptic Motor Adaptation}

Recent advances in machine learning and theoretical neuroscience have led to the ability to optimize neuroscience-derived three-factor learning rules with backpropagation through time \cite{schmidgall2021spikepropamine, schmidgall2022learning, schmidgall2023metaspikepropamine}, making powerful gradient-descent based approaches accessible for the optimization of local learning rules. These algorithms can be meta-trained through a bi-level optimization to adapt the underlying behavior of the network toward an objective \textit{during deployment} (inner-loop) via gradient descent of an objective function \textit{after deployment} (outer-loop). We extend these ideas toward the development of a motor adaptation algorithm whereby the synaptic weights of the network change based on a meta-optimized three-factor learning rule to adapt in real-time to environmental conditions which we call Synaptic Motor Adaptation (SMA).

\subsection*{A three-factor synaptic motor adaptation rule}

In MA algorithms, the role of the factor encoding module $\mu$ (Equation \ref{eq:FactorEncoder}) is to provide a \textit{context signal} for the robot so it can adapt its behavior to better suited for its environment which is constantly changing, such as walking on uneven surfaces, the existence of limb damage, or when the ground becomes slippery. This context signal changes the behavior of the robot by providing a learned embedding from the factor encoder as \textit{input} to another policy network. While this elegantly allows the robot to adapt to new environmental challenges, the \textit{fundamental behavior} of the policy is not capable of changing (i.e. the synaptic weights), rather just the information the robot has about the environment is constantly being re-estimated (its state input). This prevents the policy from actually \textit{learning} from new experience, instead, it can only update its state input based on the time history of events. 

Like other motor adaptation algorithms, SMA consists of a base policy $\pi$ which takes in robot sensory information, $x_t$, and an environment factor encoder $\mu$ which takes in privileged information $e_t$. However, SMA differs from other MA algorithms because it uses the environment factor encoder $\mu$ to produce a neuromodulatory \textit{learning signal} $m(t)$ (in our model $m_{-}(t)$ and $m_{+}(t)$) which dictates the degree with which connections are updated. This can be explained by the following equations:

\begin{equation}
    m_{+}(t), m_{-}(t) = \mu(e(t))
\end{equation}
\begin{equation}
    a(t) = \pi(x(t), a(t-1), \textbf{\textit{W}}(t), m_{+}(t), m_{-}(t)).
\vspace{2mm}
\end{equation}

In this equation, instead of a time-varying adaptive signal $z(t)$ being produced by $\mu$ there are two modulatory signals $m_{+}(t), m_{-}(t)$, and instead of $z(t)$ being given as input to $\pi$ there is a time-dependent weight parameter \textbf{\textit{W}}\textit{(t)}. The adaptive weight parameter \textbf{\textit{W}}\textit{(t)} is updated by the following equations:

\begin{equation}\label{eq:ModEligibilitySMAUpdate}
\textbf{\textit{W}}_{i,j}^{(l)}(t+\Delta\tau) = \textbf{\textit{W}}_{i,j}^{(l)}(t) + \alpha(t) \Delta_{\textbf{\textit{W}}}(t)
\end{equation}
\begin{equation}\label{eq:ModEligibilitySMADelta}
\Delta_{\textbf{\textit{W}}}(t) = m_{+,i}(t) \textbf{\textit{E}}_{+,i,j}^{(l)}(t) + m_{-,j}(t) \textbf{\textit{E}}_{-,i,j}^{(l)}(t).
\vspace{2mm}
\end{equation}

We note here that unlike in Equation 8, there are \textit{two} eligibility traces, one for the LTP dynamics $\textbf{\textit{E}}_{+,i,j}^{(l)}(t)$ + $m_{+,j}(t)$ and another for the LTD dynamics $\textbf{\textit{E}}_{-,i,j}^{(l)}(t)$. This necessitates the incorporation of two modulatory signals ($m_{+}(t)$ and $m_{-}(t)$), one for each of the eligibility traces. We see that Equations \ref{eq:ModEligibilitySMAUpdate} and \ref{eq:ModEligibilitySMADelta} update the weights of $\pi$ using the modulatory dynamics $m_{+}(t) m_{-}(t)$ produced by the environment factor encoder. That is to say, instead of determining how privileged information can best inform the network at the sensory level like traditional MA algorithms, SMA determines how to utilize privileged information to best \textit{update the base policy} $\pi$ \textit{synaptic weights}. This is possible since recent work enabled the dynamics of the three-factor learning rule to be differentiated through in spiking neural networks \cite{schmidgall2021spikepropamine, schmidgall2022learning}, and thus the policy gradient loss is backpropagated through the plasticity dynamics to optimize the modulatory signals $m_{+}(t), m_{-}(t)$ given privileged information $e(t)$. Furthermore, the modulatory signal dynamics $m_{+}(t), m_{-}(t)$ are approximated by an environment factor estimator $\phi$, enabling the learned adaptive dynamics to be utilized without privileged information.

Once the weight delta $\Delta_{\textbf{\textit{W}}}(t)$ has been computed via the eligibility and modulatory trace dynamics, it is multiplied by a time-varying term $\eta(t)=\text{exp}(1/t)-1$ before being incorporated into the synaptic weights. We refer to this term as the stabilization variable, and it exponentially decays to zero as $t \to \infty$ in order to stabilize the weight dynamics as the quadruped adapts to its environment conditions. We found that without this term, the weight dynamics are unstable across time horizons greater than what the network was trained for and, with the addition of the stabilization term, sustained control of the quadruped can be maintained over long time horizons.





\begin{table*}[]
\begin{tabular}{@{}lcccccc|c@{}}
\toprule
 & No noise & Rough terrain & Motor gain & P-gain & D-gain & Friction  \\ \midrule
Non-Adaptive SNN & 7.4 & 4.6  & 4.5 & 6.1 & 6.4 & 7.0  \\
Plastic SNN & 7.2 & 4.5  & 4.2 & 6.5 & 6.7 & 7.2  \\
RMA & 8.2 & 5.7  & 5.1 & 6.8 & 7.3 & 7.6  \\
\textbf{SMA} & 8.1 & 5.9  & 5.7 & 6.9 & 7.1 & 7.7  \\ \midrule
RMA Expert & 8.5 & 5.9 & 5.4 & 7.2 & 7.7 & 8.0 \\
\textbf{SMA Expert} & 8.2 & 6.2 & 6.1 & 7.5 & 7.5 & 7.9  \\ \midrule
** Noise Range &  &  $V_{scale}$=0.25, $H_{scale}$=0.8 & [0.8, 1.2] & [12.5, 37.5] & [0.25, 0.75] & [0.1, 2.75]  \\ \bottomrule
\end{tabular}
\caption{Simulation testing results. Each entry is defined as $\sum_{i} R_{i} \cdot P_{i}$ where the total sum of rewards for a rollout is $R_{i}$ and the probability of the domain randomization sample is $P_{i}$. $P_{i}$'s are sampled at discrete intervals along the noise ranges listed above. SMA and RMA experts are the SMA and RMA models being provided with privileged information $e(t)$ together with their extrinsics encoding module $\mu(t)$. The non-experts are using approximations.}
\end{table*}


\section*{Experimental setup}

\subsection*{Parallel reinforcement learning}

We use a modified implementation of the Proximal Policy Optimization (PPO) algorithm \cite{schulman2017proximal} specifically designed for massively parallelized reinforcement learning \cite{rudin2022learning} on the GPU. This algorithm allows learning from thousands of robots in parallel with minimal algorithmic adjustments.

The batch size, $B=n_{steps}n_{robots}$, is a critical hyper-parameter for successful learning in on-policy algorithms such as PPO. If the batch size is too small, the algorithm will not learn effectively, while if it is too large, the samples become repetitive, leading to wasted simulation time and slower training. To optimize training times, a small $n_{steps}$ must be chosen, where $n_{steps}$ is the number of steps each robot takes per policy update, and $n_{robots}$ is the number of robots simulated in parallel. The algorithm requires trajectories with coherent temporal information to learn effectively, and the Generalized Advantage Estimation (GAE) \cite{schulman2015high} requires rewards from multiple time steps to be effective. In previous work \cite{rudin2022learning}, a minimum of 25 consecutive steps or 0.5 s of simulated time is demonstrated to be sufficient for the algorithm to converge effectively. It is shown that using mini-batches of tens of thousands of samples can stabilize the learning process without increasing the total training time for massively parallel use cases.

During the training of the PPO algorithm, robots need to be reset periodically to encourage exploration of new trajectories and terrains. However, resets based on time-outs can lead to inferior critic performance if not handled carefully. These resets break the infinite horizon assumption made by the critic, which predicts an infinite horizon sum of future discounted rewards. To address this issue, like in \cite{rudin2022learning}, the environment interface is modified to detect time-outs and implement a bootstrapping solution that maintains the infinite horizon assumption. This approach mitigates the negative impact of resets on critic performance and overall learning, as demonstrated through its effect on the total reward and critic loss.

The handling of resets must also take into account the temporal dynamics of the synaptic state variables, e.g. the eligibility and synaptic traces. When working with PPO, which iteratively recalculates log probabilities from old data, this is not necessarily trivial. The challenge lies in the PPO algorithm's non-temporal treatment of data, where typically minibatches randomly sample \textit{states} at arbitrary points in \textit{time}. This is a challenge because as a temporally-dependent policy changes across time (e.g. recurrent networks, plastic networks), unlike non-temporal ANNs, the dynamic equation that led to an action $a(t)$ at time \textit{t} was dependent on all timesteps \textit{0 $\leq \tau$ < t}. Thus, to calculate $a(t)$, PPO must be modified to incorporate \textit{rollout} mini-batches where, instead of randomly sampling points in time for evaluation, entire robotic trajectories are randomly sampled and the dynamic equations (e.g. synaptic weights) are rolled out in time. In other words, since $B=n_{steps}n_{robots}$, minibatches sample along $n_{steps}$ but since there is temporal dynamics and the entire $n_{steps}$ must be rolled out we sample along $n_{robots}$.


\subsection*{Observations, actions, and noise}

Base linear and angular velocities, measurement of the gravity vector, velocity commands, joint positions and velocities, and the previous actions taken by the policy. Each of these values are scaled by a constant factor \textit{(see Appendix)}. Additionally, random noise is added to the sensor readings sampled from the following uniform distributions:

\begin{enumerate}
    \item Joint positions: $\pm 0.01$ rad
    \item Joint velocities: $\pm 1.5$ rad/s
    \item Projected gravity: $\pm 0.05$ m/$s^{2}$
    \item Base linear velocities: $\pm 0.1$ m/s
    \item Base angular velocities: $\pm 0.2$ rad/s
\end{enumerate}

Observation noise is added to account for the inherent variability in the environment, such as sensor noise and measurement errors. Introducing noise to the observations helps the policy learn to be robust to variations, improves its ability to generalize to new situations, and otherwise better benchmarks adaptivity of the controller. 

The action $a(t)$ taken by the policy $\pi$ is a desired joint position which is sent to a PD controller to calculate torques for the joints of the robot via the following equation:
\begin{equation}
\tau(t) = K_p(c_{a}a(t) + q_0 - q(t)) - K_d\dot{q}(t)
\end{equation}

where: $\tau(t)$ is the torque output at time $t$. $q(t)$ and $\dot{q}(t)$ are the current joint position and velocity, respectively. $q_0$ is the default joint position. $c_{a}a(t)$ is the scaled action at time $t$, with $c_{a}$ being the action scale factor. $K_p$ and $K_d$ are the PD gains, which are optimized by the experimenter as a hyperparameter. For our experiments we chose $K_p=20$ and $K_d=0.5$. Actions are further scaled by a constant $0.25$ to account for a physics simulator decimation size of four (simulation updates per policy update). Directly outputting torques is also an option, but we found outputting a target position into a PD controller provides quicker learning and smoother gaits.

\subsection*{Reward Terms}

The reward function reinforces the robot to follow a velocity command along the \textit{x, y,} and angular ($\omega$) axes and penalizes inefficient and unnatural motions. The total reward is a weighted stum of nine terms detailed below. To create smoother motions we penalize joint torques, joint acceleartions, joint target changes, and collisions. Additionally, there is a reward term to encourage taking longer steps which produces a more visually appealing set of behaviors.

\begin{enumerate}
    \item Tracking forward velocity: $\phi(\textbf{v}^{*}_{b, xyz} - \textbf{v}_{b, xyz})$
    \item Tracking angular velocity: $\phi(\boldsymbol{\omega}^{*}_{b, z} - \boldsymbol{\omega}_{b, z})$
    \item Angular velocity penalty: -$||\boldsymbol{\omega}^{*}_{b, xy}||^{2}$
    \item Torque penalty: -$||\boldsymbol{\tau}_{j}||^{2}$
    \item DOF Acceleration: -$||\ddot{\textbf{q}}_{j}||^{2}$
    \item Action rate penalty: -$||\textbf{q}^{*}_{j}||^{2}$
    \item Collision: -$\textit{n}_{collisions}$
    \item Feet air time: $\sum_{f=0}^{4}(\textbf{t}_{air,f}-0.5)$
\end{enumerate}

In these equations we define $\phi(x) = \textit{exp}(-\frac{||x||^{2}}{0.25})$. Values $\textbf{v}^{*}_{b, xyz}$, $\boldsymbol{\omega}^{*}_{b, z}$, and $\boldsymbol{\omega}^{*}_{b, xy}$ are superscripted with $*$ to represent that they are the target command, with the non-superscripted value as the true value. Finally, the total sum of reward terms $r(t) = \sum_{i}r_{i}(t)$ at each timestep $t$ is clipped to be a positive value. This requires more careful reward tuning upon initialization, but prevents the robot from finding self-terminating solutions.

\subsection*{Pre-training an SNN}

Instead of training the SMA network entirely from scratch, which takes significant compute resources, we initially train a non-plastic SNN without any noise in the simulation to act as the foundation. Once this network is fully trained, plasticity is added to the third layer of the policy network and noise is added to the simulator, from which the network is optimized as is outlined in the section \textit{Synaptic Motor Adaptation}.

\subsection*{Results and analysis}

We report the performance measurements of several models including: a non-plastic SNN (fixed-weights), a plastic SNN without SMA, RMA without adaptation, RMA with adaptation, and SMA. Additionally, the performance of the RMA and SMA experts are recorded, with an expert being defined as the motor adaptation algorithm provided with exact extrinsics information as defined in \cite{kumar2021rma} rather than its embedding approximation (see Equation \ref{eq:embedapprox}). The performance measurements in Table 1 are defined as follows: $\sum_{i} R_{i} \cdot P_{i}$ where the total sum of rewards for a rollout is $R_{i}$ and the probability of the domain randomization sample is $P_{i}$, with $P_{i}$'s sampled at discrete intervals along the noise ranges listed at the bottom of Table 1.

\paragraph{Adaptation to noise} The discrepancies between the physics simulator and real hardware are what lead to difficulties translating models trained in simulation to real robots. Recent ideas in robotic learning have led to the belief that adding significant "domain noise," which is noise added to the environment during training to change the physical dynamics (e.g. contact dynamics and friction), could prevent simulation overfitting and lead to a policy that can provide control in a variety of physical conditions. However, these methods tend to provide \textit{robust} policies that have unsophisticated and jerky movements. Thus, recent efforts have gone toward developing policies that \textit{adapt} to domain noise, fine tuning their control with respect to noise instead of simply becoming robust to all forms of noise.

Both the motor gain and P-gain were areas in which the SMA policy demonstrated improvements in adaptation compared with an RMA policy, with the D-gain and friction not being too far behind. However, D-gain and friction noise were not demonstrated to outperform RMA, but were close in performance. Compared with the three non-MA algorithms, there are clear benefits in performance compared with MA algorithms. However, between RMA and SMA, the performance difference is relatively small, even among the tasks that SMA obtains higher performance. This could suggest one of several things: (1) RMA and SMA are approaching a performance upper bound on adaptivity given the defined degree of noise or (2) these algorithms approached similar performance and there is more progress to be made. However, more experimentation is needed to determine this.

\paragraph{Adaptation to terrain} The ability to adapt to novel forms of terrain that were outside the scope of training is a crucial capability required for legged robots. This is because the complexity of the real world cannot adequately be captured by simulated environments, and thus, in addition to model-derived forms of noise, adaptation to terrain must be demonstrated for a motor adaptation algorithm that will be useful in the real world. For the introduction of rough terrain in our work we used \textit{perlin} fractal noise. Perlin noise generates natural-looking noise by defining the slope of the noise function at regular intervals, creating peaks and valleys at irregular intervals instead of defining the value of the noise function at regular intervals. Perlin fractal noise is a type of perlin noise that uses multiple octaves (layers) of noise to create a more complex and varied pattern. Each octave is a version of the perlin noise function with a different frequency and amplitude, and the outputs of each octave are combined together to create a final noise pattern. By adjusting the frequency, amplitude, and number of octaves used, the resulting noise can range from smooth and gentle to rough and jagged, making it useful for generating natural-looking textures and terrain. The parameters for the fractal noise are as follows: number of octaves = 2, fractal lacunarity = 2.5, fractal gain = 1.5, fractal amplitude = 1, vertical scale = 0.35, and horizontal scale = 0.08.

As is demonstrated in Figure 2, robots are trained to produce locomotion entirely on flat terrain. Unlike the analysis of adaptation to noise (e.g. motor strength noise), terrain noise is not explicitly encountered during training. Adaptation to rough terrain was among the least transferable skill from the algorithms without motor adaptation, with the non-adaptive SNN, plastic SNN, and RMA without adaptation failing to demonstrate clear generalization to the rough terrain domain. However, both the RMA and the SMA trained robots were successfully able to walk across rough terrain (without falling) despite being trained entirely on flat terrain. Interestingly, this is in spite of terrain and foothold data not being provided to the MA algorithms as privileged information.

\section*{Discussion}

We presented the SMA algorithm for real-time adaptation of a quadrupedal robot toward changes in motor strength, P and D gains, friction coefficients, and rough terrain using three-factor learning. This algorithm was compared to the state-of-the-art motor adaptation algorithm RMA \cite{kumar2021rma} and was demonstrated to perform similarly or better on motor control problems that required real-time adaptation. While adaptation improvements are relatively modest compared to the RMA algorithm, we expect further improvements with using more dynamically rich plasticity rules (e.g. triplet, voltage-based), neuron models (e.g. adaptive, resonate-and-fire), propagation delays, surrogate gradient techniques, and modulatory dynamics. Another potential direction is toward developing methods of synaptogenesis, such that the value of the weights along with the network connectivity mapping is learned. Previous methods have incorporated synaptogenesis through genetic algorithms \cite{manngaard2018structural}, neural cellular automata \cite{mordvintsev2020growing, najarro2022hypernca, gilpin2019cellular}, and online random mutations \cite{schmidgall2021self}--a solution utilizing backpropagation has yet to be developed. Much further work aims to enable learning completely novel behaviors (e.g. vaulting) purely through meta-optimized three-factor learning.

There are two clear directions that this work intends building toward: (1) using three-factor learning to transfer from simulation to real hardware and (2) exemplifying this algorithm on neuromorphic hardware. While the path toward transferring from simulation to hardware is clear, further advancements toward the optimization of plasticity rules is required before utilizing current neuromorphic systems. This is because many current neuromorphic systems (1) have propagation delays which are not incorporated into the plasticity dynamics of this work, and (2) are heavily numerically quantized whereas this work was built on fixed-point math. While much of the work toward differentiating through these dynamics has already been solved \cite{schmidgall2021spikepropamine}, meta-optimizing three-factor learning rules through these dynamics is a less explored direction (\textit{see} \cite{schmidgall2023metaspikepropamine}).

A primary limitation to our approach lies in the addition of the stabilization term $\alpha(t)$ in Equation \ref{eq:ModEligibilitySMAUpdate}. This stabilization term allows for rapid weight modifications at the beginning of the episode as the quadruped learns from interacting with the environment, and then exponentially decays its effect over time to consolidate the weights. This decay is important for long-term adaptation, particularly for an additive pair-based STDP rule which is not temporally stable \cite{kepecs2002spike}. While the neuromodulatory dynamics are capable of modulating these changes, we found that without the stabilization term, the weights still tend to diverge into bimodal distributions. This is potentially an effect of truncating the gradient in time, which does not allow proper credit assignment caused by temporally distant modifications. Future work will aim to determine \textit{when} weight modifications should occur and \textit{which} synapses should be modified in the meta-optimization dynamics.

Overall, this work introduces an exciting path toward rapid adaptation on robotic systems using neuroscience-derived models of three-factor learning and we hope it inspires further applications of three-factor learning on robotic systems.

\bibliographystyle{naturemag}
\bibliography{sample-base}

\section*{Appendix}

\subsection*{Training Hyperparameters}

\subsubsection*{Base policy hyperparameters (PPO)}
\begin{enumerate}
    \item Batch size: 250,000 (25*10000)
    \item Entropy coefficient: 0.01
    \item Discount factor: 0.99
    \item GAE discount factor: 0.95
    \item PPO Epochs: 5
    \item PPO Clip: 0.2
    \item Minibatches: 4
    \item Initial learning rate: 1e-3
    \item Learning rate decay: 0.999
    \item Maximum grad norm: 1.0
    \item Gradient steps: 2000
\end{enumerate}

\subsubsection*{Neuron, network, and plasticity hyperparameters}
\begin{enumerate}
    \item LIF time constant: exp(-$\frac{1}{10}$)
    \item Initial STDP trace constant: exp(-$\frac{1}{10}$)
    \item Initial eligibility trace time constant: exp(-$\frac{1}{200}$)
    \item Plastic weights update scale: 1e-3
    \item Network hidden dimensions: 512, 128, 64 
    \item LIF firing threshold: 1.0
    \item Weight initialization ranges: 
    \item Initial trace learning rate: $lr_{i,j} \sim$ U(0, 1) (lr * 1e-3)
    \item Learning rate decay: 0.995
\end{enumerate}

\subsubsection*{SMA policy hyperparameters (A2C)}
\begin{enumerate}    
    \item Batch size: 61440 (30*2048)
    \item Entropy coefficient: 0.005
    \item Discount factor: 0.99
    \item GAE discount factor: 0.95
    \item Initial learning rate: 3e-4
    \item Learning rate decay: 0.999
    \item Maximum grad norm: 1.0
    \item Gradient steps: 5000
    \item BPTT truncation window: 30
    \item Synaptic Trace Penalty : 1e-2
\end{enumerate}

\subsubsection*{ROA policy hyperparameters (A2C)}
\begin{enumerate}    
    \item Batch size: 61440 (30*2048)
    \item Entropy coefficient: 0.005
    \item Discount factor: 0.99
    \item GAE discount factor: 0.95
    \item Initial learning rate: 3e-4
    \item Learning rate decay: 0.999
    \item Maximum grad norm: 1.0
    \item Gradient steps: 5000
    \item ROA state history length: 10
    \item Embedding dimensionality: 8
\end{enumerate}

\subsubsection*{Action and observation scaling}

\begin{enumerate}
    \item Joint positions: 1.0 rad
    \item Joint velocities: 0.05 rad/s
    \item Base linear velocities: 2.0 m/s
    \item Base angular velocities: 0.25 rad/s
\end{enumerate}

Observations are clipped between [-100, 100]. Similarly action \textit{torques} are clipped between torque limits, which are defined by the robot manufacturer.

\subsubsection*{Reward scaling}

\begin{enumerate}
    \item Tracking forward velocity: 1.0
    \item Tracking angular velocity: 0.5
    \item Angular velocity penalty: -0.05
    \item Torque penalty: -0.0002
    \item DOF Acceleration: -2.5e-7
    \item Action rate penalty: -0.01
    \item Collision: -1.0
    \item Feet air time: 1.0
\end{enumerate}

\subsubsection*{Details on ROA}

The motor adaptation algorithm presented in the section \textit{Motor Adaptation Algorithms} was Rapid Motor Adaptation (RMA) \cite{kumar2021rma}. Improvements to this algorithm were realized by understanding that there is an information gap between the full state available to the environment factor encoder $\mu$ and the environment factor estimator $\phi$ in a follow on work \cite{fu2023deep}. Due to this, the factor encoder may generate an embedding that is not possible for the estimator to predict based on its current information, and hence there is a regression gap. To overcome this, a penalty can be added to the RMA training equations as follows:

\begin{equation*}
    L(\theta_{\pi}, \theta_{\mu}, \theta_{\phi}) = -J(\theta_{\pi}, \theta_{\mu}) + \lambda ||z^{\mu} - \text{sg}[z^{\phi}]|_{2} + ||\text{sg}[z^{\mu}] - z^{\phi} ||_{2}
\end{equation*}

which is a resulting algorithm known as \textit{Regularized Online Adaptation (ROA)}. Here, the encoder and decoder are trained jointly, where the adaptation module is training by imitating $z^{\mu}$ online and $z^{\mu}$ is regularized to avoid large deviations from the embedding estimate $z^{\phi}$. Future advancements in SMA should include this regularization to learn a signal that can be better represented by local information.

\subsubsection*{Hardware Resources}
\begin{itemize}
    \item Graphic Processing Unit (GPU): NVIDIA Quadro RTX 8000 (48 GB)
    \item Processor: Intel Xeon(R) CPU E5-2623 v3 @ 3.00GHz × 16 
    \item Memory: 252 GB
    \item Operating System: Ubuntu 20 LTS
\end{itemize}

\end{document}